\documentclass[10pt, a4paper]{article}

\usepackage[]{lrec2026} % this is the new style
\usepackage{graphicx} %Daniela
\usepackage{subcaption} %Daniela
\usepackage{paralist} %Daniela
\usepackage{mathtools}
\usepackage{makecell} %Daniela

\title{Linguistic Distancing on Social Media: Indicators of Emotion Regulation Across Age Groups}

\name{Daniela Teodorescu$^{\alpha*}$, Saif M. Mohammad$^\dagger$, Alona Fyshe$^{\alpha * \ddagger}$} 

\address{$^\alpha$Alberta Machine Intelligence Institute, $^*$Department of Computing Science,
University of Alberta\\
$^\dagger$National Research Council Canada, \\
$^\ddagger$Department of Psychology, University of Alberta\\
         \{dteodore,alona\}@ualberta.ca, saif.mohammad@nrc-cnrc.gc.ca}
\abstract{
Managing our emotional responses to events is key to emotional well-being, a process referred to as emotion regulation in psychology. 
Previous work has established that the degree to which we distance events is a type of emotion regulation. 
When we psychologically distance from events there can be markers in our language. These markers have been referred to as linguistic distancing. 
We build upon a previous metric to operationalize linguistic distancing, and explore how it changes across the lifespan. We explore this systematically by analyzing large amounts of social media text, a venue where people express their emotions.
By investigating how distancing varies across age groups we can better understand how emotion regulation varies with age and provide initial benchmarks on social media data. 
We provide additional evidence further strengthening the hypothesis that 
linguistic distancing 
occurs in proportionally more instances with age.
These findings align with past work in psychology which indicate improved well-being with older age.
Better understanding how linguistic distancing changes with age is important because it functions as a marker of well-being and can inform effective health interventions.
We provide a foundation for further exploring emotion regulation through linguistic distancing in text data. 
 \\ \newline  }

\begin{document}

\maketitleabstract

\section{Introduction}

%%%%%%%%%%%%%%%%%%%%%%%%%%%%%%%%%%%%%%%%%%%%%%%%%%%

%%%%%[Paragraph 1: Emotion regulation important to well-being]

Our everyday emotional experiences are not static but rather dynamic and our emotions are constantly changing over time. The way in which our emotional experiences change over time 
% due to emotional events and how we appraise them 
creates an emotional trajectory, or what some refer to as an \textit{emotion arc} \cite{mohammad-2011-upon, reagan}.
% Emotion regulation aims to modify the emotional trajectory. 
Emotion regulation includes the processes by which ``we influence which emotions we have, when we have them, and how we experience and express them'' \cite{gross1998}.
More simply put, emotion regulation is the way in which we manage our emotions.

Emotion regulation is key to overall well-being; dysregulation of emotions is associated with psychopathologies such as major depressive disorder and social anxiety disorder \cite{aldao2010}.
There are various ways (or strategies) of regulating emotions and some of these are seen as \textit{maladaptive}. While such strategies provide momentary relief from distress, they are not effective in the long term. Maladaptive strategies include rumination (re-thinking about memories or negative experiences), avoidance (of a situation or emotional experience), and suppression (of emotions or thoughts), and each of these has been shown to be significantly associated with anxiety, eating disorders, and substance abuse \cite{aldao2010}.
On the other hand, some strategies are considered \textit{adaptive} and are seen as healthy ways of managing emotions, e.g., reappraisal (changing how one thinks about a situation to change emotional reactions), problem solving (taking actions to solve a problem), and acceptance (non-judgmental acceptance of emotions). Adaptive strategies are seen as aiding against psychopathology and use of such strategies are 
% significantly negatively 
inversely associated with psychopathology \cite{aldao2010}.
% association between their use and psychopathology.

%%%%%%%%%%%%%%%%%%%%%%%%%%%%%%%%%%%%%%%%%%%%%%%%%%%

%%%%%[Paragraph 2: age emotion regulation consequences, understanding how changes with age is important]

Given the importance of emotion regulation to well-being, understanding how it changes with age is important for providing appropriate support and interventions. 
% Emotion regulation is seen as a goal, and goals changes with various periods of life and age.
The specific emotion regulation strategies used 
% has been found to 
change throughout childhood, adolescence, young adulthood, and adulthood \cite{Nook2020Age}. 
For example, a shift has been found from emotion regulation strategies which deal with a situation behaviorally (e.g., escaping a situation), to more cognitive strategies such as seeking information \cite{Brown1991,Altshuler1995}.
% (e.g., seeking information;  Brown, Covell, & Abramovitch, 1991)
Although there are some conflicting findings on how exactly various strategies track with age, many studies point to older ages having ``enhanced emotion regulation'' \cite{gross2010}, being better at regulating their emotions \cite{Gross1997, Mehlsen2024}, and have better overall well-being than younger adults despite losses physically and cognitively \cite{gross2010,Stawski2008}. 

% talk about specific age findings

%%%%%%%%%%%%%%%%%%%%%%%%%%%%%%%%%%%%%%%%%%%%%%%%%%%

%%%%%[Paragraph 3:emotion regulation can be done in many ways, we focus on linguistic distancing, we explore linguistic distancing]

Another lens through which emotion regulation has been studied is psychological distancing.
By creating distance from an event, one can better handle their emotions related to the event and view the event objectively (allowing for a reappraisal of the event). % \cite{}
Adopting a distanced perspective has been shown to down-regulate negative affect \cite{Kross2017}, and it is a common technique used in Cognitive-Behavioral Therapy \cite{beck2020cognitive} and Dialectical Behavior Therapy \cite{linehan1993skills}. 
When psychologically distancing from an event, one is viewing the event from a third-person perspective, and often psychological distancing appears in language as markers.
For example, when distancing, there are less first-person pronouns and more past and future tense verbs 
% are used
rather than present tense verbs. These markers are described via \textit{linguistic distancing} \cite{Nook2017}. Linguistic distancing has been shown to be associated with successful emotion regulation and reduced negative affect \cite{Nook2017}. Further, in client-therapist transcripts, less linguistic distancing was associated with worse internalizing symptoms (inward focused behaviors often occurring with anxiety and depression) \cite{Nook2022}.
% talk about bidirectional relationship?
%%%%%%%%%%%%%%%%%%%%%%%%%%%%%%%%%%%%%%%%%%%%%%%%%%%

%[Paragraph 4: Linguistic distancing on social media has not been explored]
% talk about age and social media
% systematic, large scale way , we do this

Through linguistic distancing we can study emotion regulation in text; although so far only client-therapist transcripts have been analyzed for select age groups. 
While linguistic distancing has been explored across childhood, adolescence, and young adults \cite{Nook2022_adolescent,Nook2020Age}, we do not know how it changes over adulthood which is what we address in our work.
While one approach to measuring linguistic distancing across ages is through longitudinal research, that is very difficult as it would require text data from the same people across all the decades. Instead, we conduct cross-sectional research, which has its own benefits, but appropriate conclusions must be drawn that are different from what could be drawn from longitudinal research. Additionally, linguistic distancing could be influenced by the era in which people grew up; e.g., those in a certain age group may use more distancing due to world events, culture, social norms, etc. than other generations. Therefore, we explore how linguistic distancing changes in \textit{contemporary} times using a platform which many of use to communicate: social media.

Social media is a platform where we frequently communicate and express our feelings. Platforms such as Reddit and X provide a space where we can freely express our thoughts at any time of day, connect with others, and possibly receive support. 
Given the important role social media plays in everyday communication, as researchers we make use of large amounts of data to analyze linguistic distancing in a systematic way across adulthood.
% it is unknown to what extent linguistic distancing occurs in social media text and what it looks like. 
We explore how linguistic distancing varies across age groups on social media.
 We make use of a social media dataset annotated with the age of the author at the time of posting to answer the following research questions: 

\begin{itemize}
    \item How does linguistic distancing vary across age groups?\\[2pt]
    % \subitem 
    \noindent
    We hypothesize that linguistic distancing increases with age as people become better at regulating emotions based on literature in psychology. 
    % We verify if this is the case in social media posts as well.
    \vspace{-2mm}
    \item How do the various \textit{dimensions} of linguistic distancing (i.e., social distancing, temporal distancing, passive voice and abstractness) vary across age groups? Do some dimensions change more than others across adulthood?
\end{itemize}

%%%%%%%%%%%%%%%%%%%%%%%%%%%%%%%%%%%%%%%%%%%%%%%%%%%
%[Paragraph 5: Wrap up how work important]

By answering these questions, we provide important findings on how linguistic distancing tracks across age groups on social media. Through linguistic distancing, our findings allow us to better understand emotion regulation on a larger scale across age, enabling further study of this phenomenon in psychology and the social sciences. We make our code publicly available.\footnote{\url{https://github.com/dteodore/Age--Linguistic_Distancing} }
%%%%%%%%%%%%%%%%%%%%%%%%%%%%%%%%%%%%%%%%%%%%%%%%%%%
% Old structure
% [Age emotion important] 
% [linguistic distancing is a way to measure emotion regulation]
% [usually measured through questionnaires, way to measure in text ]
% [we improve metric and show its utility in capturing linguistic distancing across age]
% [specific RQs]
% [We contribute way to easily quantify distancing on text data in python (release a tool?), contribute to findings on age-emotion, allows study of vast phenomenon on social media \& well-being]

% Determining the extent to which we distance events to help regulate our emotions has traditionally been examined through questionnaires.

%%%%%%%%%%%%%%%%%%%%%%%%%%%%%%%%%%%%%%%%%%%%%%%%%%%

\section{Related Work}
\label{sec:related_work}

Below we describe past work examining the relationship between emotion regulation, linguistic distancing and how they change across the lifespan.
Afterwards, we describe markers of distancing in language. 
% linguistic distancing and emotion regulation in psychology. We also describe related work demonstrating the relationship between passive voice in language with distancing and using abstractness in text with distancing.

\subsection{Emotion Regulation, Linguistic Distancing \& Age}

Appropriately regulating emotions is key to mental health and well-being. Vast amounts of literature point to dysregulation of emotions being tied with psychopathology such as major depressive disorder and anxiety disorders \cite{gross1995,Sheppes2015,aldao2010}. The various ways in which we manage our emotions are related with health outcomes: adaptive strategies such as cognitive reappraisal are significantly associated with positive indicators of mental health, whereas maladaptive strategies such as suppression of emotions are significantly associated with negative indicators of mental health \cite{Hu2014}. Emotion regulation also changes throughout the lifespan. Infants have minimal to no control over their emotions, relying on their caregiver for support \cite{bowlby1969attachment}. Infants begin to develop approaches such as changing their eye gaze to focus on other objects as a way of signaling autonomous emotion regulation \cite{Derryberry1988}. As they become toddlers, they begin to understand language which opens a whole new world in terms of emotion regulation. Now, they can begin to understand instructions from their parents \cite{Thompson1991} and they can begin to express how they feel and socialize with others, learning to differentiate between positive and negative emotions \cite{gross1995}. %Malatesta & Izard, 1984; Saarni & von Salisch,1992
In adolescence, more sophisticated strategies develop, such as reappraisal as well as maladaptive strategies \cite{gross1995,hall1993nicotine,Felix1995}.  
While overall findings point to older age being associated with better overall well-being, control of emotions, and enhanced emotion regulation \cite{gross2010, Gross1997}, there are some discrepancies in exactly how emotion regulation changes with age, such as which strategies do we use more vs. less.
For example, some work found \textit{suppression} decreased across the life span \cite{John2004,DeFrance2019}, whereas others found \textit{suppression} increased with age \cite{Brummer_2014}.
\citet{DeFrance2019} found adolescents used \textit{distraction} and \textit{suppression} more than young adults, and young adults used more \textit{rumination}, whereas \citet{Sutterlin2012} found no differences in \textit{rumination} across ages.
\citet{John2004} found \textit{reappraisal} increased across the lifespan, whereas \citet{DeFrance2019} found no difference. 
Given the conflicting findings, we use social media data to examine how \textit{reappraisal} changes across the lifespan.

We specifically study reappraisal through linguistic distancing. Linguistic distancing has been studied across children, adolescents, and young adults when performing a reappraisal task, however no differences across ages was found \cite{Nook2020Age}. Although there were changes in the exact strategies of cognitive reappraisal employed across ages: \textit{changing circumstances} and \textit{distancing} increased across age; \textit{changing consequences} decreased across age; and adolescences used more \textit{ challenging reality} and less\textit{ problem-solving} compared to other ages. 
Linguistic distancing has also been studied specifically in adolescents: more linguistic distancing was associated with lower levels of hopelessness and higher levels of perceived agency, whereas greater use of linguistic distancing predicted fewer depressive symptoms in follow-ups \cite{Nook2022_adolescent}.
While these studies provide findings for ages 10--23 \cite{Nook2020Age} and 13--16 \cite{Nook2022_adolescent}, we are interested in changes across a vast range of ages i.e., 13--70's. Further, we investigate the natural use of linguistic distancing as it appears in everyday communication (e.g., social media), whereas past findings in psychology were based on self-reports and questionnaires (e.g., Emotion Regulation Questionnaire (ERQ) \cite{erq}, Cognitive Emotion Regulation Questionnaire (CERQ) \cite{CERQ}) or reappraisal tasks.
% whereas
While past work has created a corpus for detecting coping strategies in text \cite{troiano-etal-2024-dealing}, which is very intertwined with emotion regulation (coping strategies take direct action to manage distress), we study naturally occurring text (rather than text from role-playing scenarios) and are interested in this phenomenon across ages.

% Traditionally, work in psychology has studied emotion regulation via self-reports and questionnaires (e.g., Emotion Regulation Questionnaire (ERQ) \cite{erq}, Cognitive Emotion Regulation Questionnaire (CERQ) \cite{CERQ}).

\subsection{Distancing in Language}

%%%%%%%%%%%%%%%%%%%%%%%%%%%%%%%%%%%%%%%%%%%%%%%%%%%
%%%%%[Paragraph 5: how linguistic distancing measured, we improve it]
% \cite{Nook2022}
% \citet{Nook2025}

Distancing can occur through various dimensions, such as in terms of time, space and socially (Construal Level Theory \cite{Liberman2008,Trope2010}). Therefore, past operationalizations of linguistic distancing include measures of \textit{social distancing} and \textit{temporal distancing}.
Use of more first-person singular pronouns e.g., ``I'', ``me'', ``mine'' represents a more immersed perspective and less \textit{social} distancing, which means that one is performing less psychological distancing.
On the other hand, using more second- and third-person pronouns e.g., ``she'', ``they'', ``them'' represents using more \textit{socially} distanced language, and therefore one is doing more psychological distancing.
% has been pronouns as a measure of \textit{social distancing} and 
Using more present tense verbs represents less \textit{temporal distancing} and less psychological distancing,
whereas using more past- and future-tense signals more \textit{temporal} distancing and therefore more psychological distancing \cite{Nook2022}.
We use these definitions of \textit{social} and \textit{temporal} distancing in our work.
The bidirectional relationship between 
emotion regulation and linguistic signatures of psychological distancing
% psychological distancing and language use 
has been shown \cite{Nook2017}. 
Regulating emotions through psychological distancing has been associated with increased linguistic markers of social and temporal distance.
% and greater linguistic distancing has been associated with more successful emotion regulation. 
Likewise, using distanced language also regulated emotions and reduced negative affect \cite{Nook2017}. 
Self-distancing (``I'' vs. ``you'' pronouns) and social
withdrawal (indicated by more ``they'' pronouns), has been explored on Reddit for healthcare workers during the COVID-19 pandemic, however we are interested in linguistic distancing on a broader scale (i.e., for those beyond the healthcare field) and in day-to-day contexts outside of the pandemic \cite{ireland-etal-2022-tracking}. 

When distancing from an event psychologically, this creates a more abstract mental representation of the event -- leading to a high correlation between linguistic measures of psychological distancing and \textit{abstractness} \cite{Nook2025}. Measures of linguistic distancing and \textit{abstractness} have been found to be highly correlated; \textit{abstractness} increased when cognitive reappraisal was used to regulate emotions; when people distanced their language, \textit{abstractness} also increased; and lastly, increased \textit{abstractness} when regulating emotions was correlated with regulation success \cite{Nook2025}. Given the established relationship between \textit{abstractness} and linguistic distancing, we also incorporate a measure of \textit{abstractness} in our metric of linguistic distancing.

Work in psychology has also pointed to the relationship between using passive voice and psychological distancing \cite{passive_voice_distancing,eugene2020}. \textit{Passive voice} is where the object is placed before the verb such that the subject receives the action rather than performing it (e.g., passive voice: ``The ball was thrown by the child'' vs. active voice: ``The child threw the ball'').\footnote{https://www.grammarly.com/blog/grammar/passive-voice/} When reading passages in passive voice participants reported feeling temporal, hypothetical, and spatial distance from the events in the passage resulting in a more abstract mental construal \cite{eugene2020,Trope2003}. Past work has also studied passive voice in the context of assault, where distancing removed the action from the actor and places blame on the victim \cite{Bohner2001}.
When using passive voice, the agency of the subject is taken away, creating distance from the action. 

Given the past literature demonstrating the relationship between linguistic distancing with \textit{social} distancing, \textit{temporal} distancing, \textit{abstractness}, and \textit{passive voice}, we construct a measure of linguistic distancing which includes all four dimensions.
Analyzing how each of these dimensions changes across adulthood is informative for understanding behaviors and tendencies which is of interest not only to psychologists, linguists, and sociologists, but is also linked with mental health.
% -- a growing field in psychology -- 
% https://www.pnas.org/doi/abs/10.1073/pnas.0910651106 

% https://www.tandfonline.com/doi/pdf/10.1080/13607860902989661?casa_token=WC8Mw_L4SJMAAAAA:8iuv8i3lJKqVePEk8ECtxBRZg07qyyk-5iDLEravl7s6tUMcfrQ4UIJpM94unsmDWBbMkrqgSQ_B 

% https://www.tandfonline.com/doi/pdf/10.1080/13607863.2017.1396575?casa_token=odReaNtPb78AAAAA:7YqWy5IcglwbrGbjBOIEalBGKGei5SXZ4x7h-JdA2mWpAyqMYEmLUZXVMCnwB87vQrTWEuS-R3mT

% https://academic.oup.com/psychsocgerontology/article-abstract/59/6/P261/589694?redirectedFrom=PDF

% https://www.cambridge.org/core/services/aop-cambridge-core/content/view/4964A0661C08EB3B1313D9519DFCBFFC/S1352465813000453a.pdf/div-class-title-the-influence-of-age-on-emotion-regulation-strategies-and-psychological-distress-div.pdf

% https://www.sciencedirect.com/science/article/pii/S0191886911002881?casa_token=9uWmK5NStkwAAAAA:osROi1xWcWBtrdxLnHhIRRe9nWxx9KVJEYNV8Jnz0H_-ncdwWrcUXlAsUhgRz7UvuC_8dwfBAw

% https://journals.sagepub.com/doi/pdf/10.1177/0963721410388395?casa_token=88Sx83JxXWsAAAAA:UPnc1NIocwdn6Si7KmnNESJ-lwwXfpuG-1NTpkgscW76MCX1bKcG75oF9YAiiTVs2_j-4Zh5Gvnv

% https://www.ovid.com/journals/scjop/fulltext/10.1111/sjop.12970~the-effect-of-age-on-emotion-regulation-patterns-in-daily
% \noindent\textbf{Abstractness:}
% Using more abstract terms 

% \noindent\textbf{Passive Voice:} Past work has found that passive voice is associated with distancing of events \cite{}. in 

% "agentless actions serve to distance the writer or speaker from the text"

\begin{table}
\centering
\begin{tabular}{lrr}
\hline
\textbf{Age Group} & \multicolumn{2}{c}{\textbf{\#Posts}}\\
\cline{2-3}
 & TUSC-City & Reddit \\  %Country &
\hline
% 13--19 & 551 & 94,857 & 9,281,055\\
% 20--29 & 3,076 & 406,027 & 15,455,426\\
% 30--39 & 3,001 & 462,238 & 6,151,757\\
% 40--49 & 1,808 & 360,946 & 1,220,498\\
% 50--59 & 1,541 & 278,809 & 449,230\\
% 60--69 & 1,551 & 218,008 & 246,398\\
% 70--79 & 419 & 97,432 & 160,465\\
% 13--19  & 94,857 & 9,281,055\\
% 20--29  & 406,027 & 15,455,426\\
% 30--39  & 462,238 & 6,151,757\\
% 40--49  & 360,946 & 1,220,498\\
% 50--59  & 278,809 & 449,230\\
% 60--69  & 218,008 & 246,398\\
% 70--79  & 97,432 & 160,465\\

%New
13--19 & 94,857 & 9,281,055 \\
20--29 & 406,027 & 15,455,426 \\
30--39 & 462,238 & 6,151,757 \\
40--49 & 360,946 & 1,220,498 \\
50--59 & 278,809 & 449,230 \\
60--69 & 218,008 & 246,398 \\
70--79 & 97,432 & 160,465 \\
\hline
\end{tabular}
\caption{The number of posts across the age groups in each subset of the \textit{AgeCorpus}. 
% ``City'' refers to TUSC-City subsets of the dataset. We use this data in our experiments.
% ``Country'' refers to TUSC-Country and 
}
\label{tab:AgeCorpus}
\end{table}

\section{Dataset: AgeCorpus}
% Age Annotated Social Media Dataset

We perform our experiments on \textit{AgeCorpus}, a social media dataset containing posts annotated with the author's age at the time of writing \cite{teodorescu2026_age}. The dataset contains posts from both Reddit and X, making it a suitable dataset for exploring linguistic distance on social media platforms. Further, there are a large number of posts in the dataset, allowing us to aggregate results per age group. The dataset was collected by identifying instances where authors self-disclosed their age e.g., ``I am X years old'' or  ``Me [20F] and my best friend [21M]''. Using various high quality pattern matching templates, the age of an author can be determined at a point in time based on the timestamp of the post (seed post). 
We show the pattern templates used in Appendix \ref{appendix:regexes}.
Using the age declared at the given timestamp in the seed post, the author's age can be determined for any other post based on the difference of the timestamp to that of the seed post's. The dataset consists of thousands of users each with hundreds of posts spanning from 2010--2022 for Reddit and 2020--2021 for X. 
We show the exact number of posts per age group in Table \ref{tab:AgeCorpus}.
The X component of the dataset consists of \textit{TUSC-Country} and \textit{TUSC-City} which differ in how they were collected. TUSC-City was collected using Twitter’s free API for American and Canadian cities during 2020--2021. Whereas TUSC-Country was collected using Twitter’s Academic API for tweets from the US and Canada during 2015--2021.
Our results on the two subsets are very similar so we report those on TUSC-City since it is the larger subset and show the results on TUSC-City in the Appendix (Section \ref{appendix:tusc_country}).
We use this dataset for our experiments described in the next section.

\begin{table*}[!ht]
\centering
{\small
\begin{tabular}
{llrrrrr}
\hline
\multicolumn{1}{l}{\textbf{Dataset}} &
\multicolumn{1}{l}{\textbf{Metric}} &
\multicolumn{1}{l}{\textbf{df1}} &
\multicolumn{1}{l}{\textbf{df2}} &
\multicolumn{1}{l}{\textbf{F-statistic}} &
\multicolumn{1}{l}{\textbf{P-value}} &
\multicolumn{1}{l}{\textbf{Effect Size (\textit{est $\omega$}\textsuperscript{2})}}\\ \hline
%One-way
% Reddit & Linguistic & 6 & 32964822 & 18001.10 & \textit{p}$<$.001 & 0.003 \\
%Welch's anova
Reddit & Linguistic & 6 & 1203317.10 & 17703.63 & \textit{p}$<$.001 & 0.003 \\
% & Temporal &  & &  & \textit{p}$<$.001 & \\
% & Social &  & &  & \textit{p}$<$.001 &  \\
% & Passive &  &  &  & \textit{p}$<$.001 &  \\
% & Abstract &  &  &  & \textit{p}$<$.001 &  \\

% TUSC-Country & Linguistic & 6 & 11940 & 29.66 & \textit{p}$<$.001 & 0.015 \\
% & Temporal & 6 & 11940 & 2.59 & \textit{p}$=$.016 & 0.001\\
% & Social & 6 & 11940 & 35.43 & \textit{p}$<$.001 & 0.017 \\
% & Passive & 6 & 11940 & 7.71 & \textit{p}$<$.001 & 0.004 \\
% & Abstract & 6 & 11940 & 10.68 & \textit{p}$<$.001 & 0.005 \\

% One-way
% TUSC-City & Linguistic & 6 & 1918310 & 2822.47 & \textit{p}$<$.001 & 0.009 \\

% Welch's anova
TUSC-City & Linguistic & 6 & 539267.68 & 2830.26 & \textit{p}$<$.001 & 0.009 \\
% & Temporal & 6 & 1918310 & 218.15 & \textit{p}$<$.001 & 0.001\\
% & Social & 6 & 1918310 & 3636.48 & \textit{p}$<$.001 & 0.011 \\
% & Passive & 6 & 1918310 & 567.58 & \textit{p}$<$.001 & 0.002 \\
% & Abstract & 6 & 1918310 & 853.84 & \textit{p}$<$.001 & 0.003 \\
 
\hline
\end{tabular}
}
% \vspace*{-1mm}
\caption{The degrees of freedom (for the numerator and denominator), F-statistic, p-value, and effect size in the one-way Welch's ANOVA test for differences in linguistic distancing between age groups. 
% Welch's ANOVA was performed for Reddit and TUSC-City. % ``emo.'' is an abbreviation for emotional.
}
\vspace*{3mm}
\label{tab:anova_distancing}
\end{table*}

\section{Experiments}

In the sections below we describe our methods for exploring how linguistic distancing varies across age groups.

\subsection{Computing Linguistic Distancing in Text}

% We compute a linguistic distancing score per post. We then average the scores per age group to analyze how linguistic distancing changes across ages. We describe each of the dimensions in our linguistic distancing score below.

We build on \citet{Nook2022}'s work on computing linguistic distancing in text. 
Specifically, the authors operationalize linguistic distancing as being composed of two components: \textit{social} distancing and \textit{temporal} distancing.
\textit{Social} distancing is measured through the proportions of 
pronouns present in the text.
Fewer first-person singular pronouns are associated with 
% more distancing.
proportionally more instances of distancing.
Here, we analyze the proportions of pronouns present in an instance or a post.
% and fewer
% first-person singular pronouns, 
More specifically, \textit{social} distancing is computed per post. 
% as: (freq(second person) + freq(first-person plural) + freq(third-person singular) + freq(third-person plural))/(freq(second person) + freq(first-person plural) + freq(third-person singular) + freq(third-person plural) + freq(first-person singular)). We use freq() to represent the frequency of a pronoun in a post. 
Let $f_{\mathrm{1p,sg}}, f_{\mathrm{1p,pl}}, f_{\mathrm{2p}}, f_{\mathrm{3p,sg}}, f_{\mathrm{3p,pl}}$ 
denote the corresponding frequencies of first-person singular, first-person plural, second-person, third-person singular and third-person plural pronouns respectively. Then the social distancing ($social$) component of linguistic distancing ($LD$) per post or instance ($t$) is computed as:
\[
LD_{social}(t) =
\frac{
f_{\mathrm{2p}} + f_{\mathrm{1p,pl}} + f_{\mathrm{3p,sg}} + f_{\mathrm{3p,pl}}
}{
f_{\mathrm{2p}} + f_{\mathrm{1p,pl}} + f_{\mathrm{3p,sg}} + f_{\mathrm{3p,pl}} + f_{\mathrm{1p,sg}}
}
\]

% \begin{equation}
%     \label{eq:1}
%   =  (second\_person + first\_person\_plural + \\
%   third\_person\_singular + third\_person\_plural)
%    / total\_pronouns
% \end{equation}

\textit{Temporal} distancing is captured through the use of verb tense.
More past and future tense verbs are associated with
% more
proportionally more instances of distancing. 
We compute \textit{temporal} distancing per post.
% is computed as: 
Let $f_{\mathrm{past}}, f_{\mathrm{present}}, f_{\mathrm{future}}$ denote the corresponding frequencies of past-tense, present-tense and future-tense verbs.
Then the temporal distancing ($temp$) component of linguistic distancing ($LD$) per post or instance ($t$) is computed as:
% (freq(past) + freq(future))/(freq(past) + freq(future) + freq(present)). 
\[
LD_{temp}(t) = 
\frac{
f_{\mathrm{past}} + f_{\mathrm{future}}
}{
f_{\mathrm{past}} + f_{\mathrm{future}} +
f_{\mathrm{present}}
}
\]
% Here, we use freq() to represent the frequency of a verb-tense in a post. 

% [Our additional metrics]
We further build on the metric for linguistic distancing by including a measure of \textit{abstractness} and \textit{passive voice}.
% We described the relationship between \textit{abstractness} and linguistic distancing in the Related Work,
We include \textit{abstractness} in our measure of linguistic distancing as more abstract mental representations of an event are linked with more successful emotion regulation, and more abstract language is highly correlated with more distanced language \cite{Nook2025}. More details on this are described in Section \ref{sec:related_work}. 
% , \citet{Nook2025} demonstrated the relationship between linguistic distancing and abstractness in language, so we also include it in our metric. 
To quantify \textit{abstractness} we use a word list with ratings of concreteness for forty-thousand words \cite{abstractness2014}.\footnote{We opt for this method of determining abstractness since \citet{Nook2025} use the LIWC Lexicon \cite{pennebaker2001linguistic}, which is not freely available.} Each word is rated on a 5-point scale from abstract to concrete which we invert to be consistent with other measures.
% so we simply inverted the scale (e.g., 1 represents abstract, 5 represents concrete so we compute abstract\_score = 6.0 - concreteness\_score).
We calculate the \textit{abstractness} score ($LD_{abs}$) per post ($t$) as the average of the abstractness score of each word at index $i$ ($abs(t_{i})$) of the instance with length $n$:
% Then the abstractness  component of linguistic distancing per post  is computed as:
\[
LD_{abs}(t) =  \frac{1}{n}
\sum_{i=1}^{n} abs(t_{i}) 
\]

Lastly, we determine whether a text contains \textit{passive voice} or not using the PassivePy tool \cite{PassivePy}. PassivePy uses POS tagging, dependency parsing and rule-based matching for sentence structures/forms in order to determine whether passive voice is present in the text or not. We integrate this binary score of passive voice ($pass(t)$) per instance ($t$) into our metric for linguistic distancing.

\[
LD_{pass}(t) = pass(t) 
\]

% Once scores for each dimension of linguistic distancing have been computed, 
Each of these measures are on different scales. Therefore, we standardize all the scores for each dimension individually such that each dimension has scores with a mean of 0 and a standard deviation of 1.\footnote{Standardizing is the same as computing the z-score.} Then we compute the linguistic distancing ($LD$) score per post ($t$) as the average of the four standardized ($stand()$) dimensions: 

% \[
% LD(t) = 
% \frac{
% LD_{temp}(t) + LD_{social}(t) + LD_{abs}(t) \\
% + LD_{pass}(t)
% }{
% 4
% }
% \]

% \begin{multline*}
%     LD(t) = \frac{ LD_{temp}(t) + LD_{social}(t) + LD_{abs}(t)
%     + LD_{pass}(t)}{4}
% \end{multline*}

\[
LD(t) = \frac{\splitfrac{stand(LD_{temp}(t)) + stand(LD_{social}(t))}{+ stand(LD_{abs}(t)) + stand(LD_{pass}(t))}}{4}
\]

\noindent\textbf{Linguistic Distancing Per Age Group:}
The next step is to aggregate scores per age groups. For our analyses we consider the following age groups: 13--19 (teens), 20--29 (twenties), 30--39 (thirties), 40--49 (forties), 50--59 (fifties), 60--69 (sixties), 70--79 (seventies).
We group instances based on the age of the author at the time of writing. Then we compute the linguistic distancing score per group as the average of all the scores of instances within the particular group.

\subsection{Statistical Analyses}

To compare whether there is a statistically significant difference in linguistic distancing measures across age groups, we perform a one-way ANOVA test. Before performing inferential tests, we ensure that our data met ANOVA's prerequisites. Namely, we tested that the data are independent of each other, each independent variable is approximately normally distributed, and each group has roughly the
same variance (homoscedasticity). 
Next, we use a qq-plot 
of the residuals 
to verify that each score is normally distributed for each group. 
In addition to the normality assumption, we can assume the estimate of the mean is normally distributed due to the large sample size (law of large numbers) according to the central limit theorem. 
% the Shapiro-Wilk Test for Normality to verify whether the residuals for each group are normally distributed. The data met this assumption.
To verify whether each group has approximately the same variance, we performed Levene's test 
% and the result was not significant for TUSC-Country,
and the result was significant ($p<0.001$)
% however was 
for Reddit and TUSC-City, 
meaning that the age groups 
% have equal variance in TUSC-Country but not in 
do not have equal variance in
the Reddit and TUSC-City subsets of the dataset. 
We can assume each data point is independent of others since each post is generally independent of one another. While it is possible that while browsing social media, a post may influence others to some degree, this is largely not the case. 
Therefore, 
% we can perform a one-way ANOVA for TUSC-Country, but 
we
will perform a one-way Welch's ANOVA for TUSC-City and Reddit. Results for Levene's tests can be found in Appendix \ref{appendix:levenes}.
After we determined our data met the assumptions, we performed the one-way ANOVA.

\begin{figure*}[!ht]
    \centering
    \begin{subfigure}[b]{0.45\textwidth}
    \centering
    % \subcaption{}
    \includegraphics[width=\textwidth]{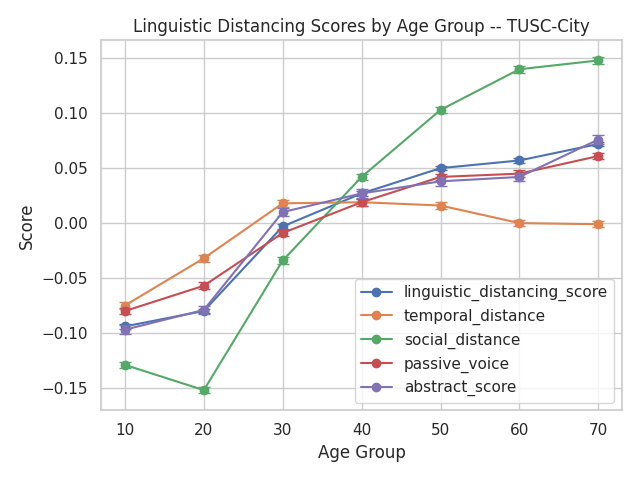}
    % \caption{}
    \label{fig:tusc_city_linguistic_dist_age}
    \end{subfigure}
    % \hfill
    % \begin{subfigure}[b]{0.45\textwidth}
    % \centering
    % % \subcaption{}
    % \includegraphics[width=\textwidth]{images/linguistic_distancing_age_country_lineplot.png}
    % % \caption{}
    % \label{fig:tusc_country_linguistic_dist_age}
    % \end{subfigure}
    %     \vspace*{-3mm}
    \begin{subfigure}[b]{0.45\textwidth}
    \centering
    % \subcaption{}
    \includegraphics[width=\textwidth]{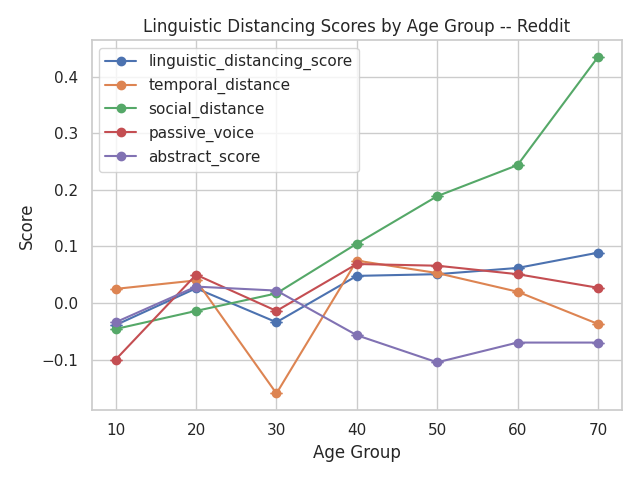}
    % \caption{}
    \label{fig:tusc_country_linguistic_dist_age}
    \end{subfigure}
        \vspace*{-3mm}        
    \caption{\textbf{Linguistic Distancing} scores for TUSC-City, and Reddit subset of the \textit{AgeCorpus} (blue line). Individual components of linguistic distancing across age groups are shown: temporal distance (orange line), social distance (green line), passive voice (red line) and abstractness (purple line). Error bars represent the standard error of the mean.}
     % TUSC-Country 
    \label{fig:tusc_linguistic_distancing_age}
\end{figure*}

\section{Results}
 
\subsection{RQ1: How Does Linguistic Distancing Vary Across Age Groups?}
In Figure \ref{fig:tusc_linguistic_distancing_age}, we show the trend of linguistic distancing across age groups. We see that regardless of the dataset (i.e., X or Reddit), linguistic distancing (blue line) 
% increases with age. By increasing, this means
% % that the overall \textit{intensity} of 
% linguistic distancing 
occurs more \textit{frequently} with age. This upwards trend is consistent across age groups (except with a dip in the 30's on Reddit). This result is also reflected in the statistical analyses. We show in Table \ref{tab:anova_distancing} the F-statistic and p-value for the ANOVA tests. 
For all datasets, there is a significant increase in linguistic distance with age
% Reddit: $F(6,32964822)=18001.10$, $p<0.001$,$\eta^2= 0.003$;
(Reddit: $F(6,1203317.10)=17703.63$, $p<0.001$,$\eta^2= 0.003$;
% TUSC-Country: $F(6,11940)=29.66$, $p<0.001$, $\eta^2= 0.015$; 
% TUSC-City: $F(6,1918310)=2822.47$, $p<0.001$, $\eta^2= 0.009$
TUSC-City: $F(6,539267.68)=2830.26$, $p<0.001$, $\eta^2= 0.009$).
The effect size on the X dataset is near 0.01 which is considered small, and the effect size on the Reddit dataset is even smaller.\footnote{Effect size of 0.01 is small; 0.06 is medium; and 0.14 is large. Effect sizes help us quantify whether differences are practically meaningful in the real-world.}

\noindent\textbf{Discussion:}
Past literature has found that overall well-being, lower negative affect, higher positive affect and improved emotion regulation skills is associated with age \cite{gross2010, Charles2023,Stone2010,Isaacowitz}. Our findings show that with age, 
% linguistic distancing 
% increases. 
there are proportionally more instances of distancing.
Given the relationship between linguistic distancing and emotion regulation, our findings support more use of reappraisal with age and in turn better well-being in the older ages. The drop in the 30's Reddit data is interesting as past work has found a dip in well-being during midlife (e.g., 35's--50's) \cite{Blanchflower2008,Blanchflower2021,Stone2010}, however, there are some theories opposing the U-shape of happiness \cite{Galambos}.

% \begin{figure}[!ht]
% \begin{center}
% \includegraphics[width=\columnwidth]{home-palacio.jpg}
% \caption{The caption of the figure.}
% \label{fig:tusc_linguistic_distancing_age}
% \end{center}
% \end{figure}

\subsection{RQ2: How does Social Distancing, Temporal Distancing, Abstractness, and Passive Voice Change with Age?}

In Figure \ref{fig:tusc_linguistic_distancing_age}, we see the trend for each of the dimensions of linguistic distancing across age groups. Every dimension (i.e., \textit{temporal} distance, \textit{social} distance, \textit{abstractness}, \textit{passive}) 
occurs more frequently
% generally 
% occurs proportionally in more instances 
% increases 
with age, although some have a steeper slope. 
\textit{Social} distancing has the steepest slope across datasets based on the graph. 
% This means that across age groups, less first-person pronouns are used.
 This means that the rate at which \textit{social} distancing increases with age is higher than the rate for other dimensions.
 In addition, \textit{social} distancing also starts off at a lower point on the y-axis than all the other dimensions, allowing more room to grow with age.
 However, on Reddit \textit{abstractness}, \textit{temporal}, and \textit{passive} scores plateau/decrease after age 40 and onwards, and likewise with \textit{temporal} distancing on X. 
% In the 40's and onward, temporal distance has the flattest slope (and sometimes negative slope) on X and Reddit. 
This could be due to the nature of the platforms and the topics discussed on each e.g., often people describe situations on Reddit to obtain advice and therefore talk in a more present-tense, less abstract manner.
Reddit also has a dip in linguistic distancing in the 30's which is primarily driven by a decrease in temporal distance. This means that individuals are using more present tense verbs.

\noindent\textbf{Discussion:}
% While some of these dimensions have not been studied directly with age, they measure concepts which past work has revealed changes with age.
While some of these dimensions have not been studied directly with age, they measure concepts that have been shown to change with age.
% For example, higher levels of \textit{social} distancing is reflected by fewer personal-pronouns. 
For example, more \textit{social} distancing is associated with fewer personal-pronouns used. 
Various literature point to the negative relationship between age and first-person singular pronouns \cite{Schwartz2013,Pang2025,Pennebaker2003}. The trends we see across X and Reddit support this pattern.
In terms of \textit{temporal} distancing, higher levels mean that fewer present tense verbs are used which is also supported by past work which points to  more future-tense and fewer past-tense verbs with age \cite{Pennebaker2003}. 
While there are few works studying \textit{passive voice} with age, older adults were found to produce fewer passive sentences than younger adults in Korean \cite{Jee2024}.
Also, older adults (i.e., 70's) had more difficulty producing and understanding abstractness on tests \cite{Albert1990}. 
For abstractness, no statistical difference was found for the ability to define abstract words across participants aged 19--89, but the use of abstract vs. concrete words in everyday language was not examined \cite{Pezzuti}.
While some work examines how each of these components vary across childhood or in terms of acquisition (e.g., linguistic distancing: \citet{Nook2020Age, Nook2022_adolescent}; passive voice: \citet{Baldie1976,Horgan_1978}), fewer work examines how they change across adulthood. 
Further, past work examining linguistic distancing does not consider the contributions of \textit{temporal} vs. \textit{social} distancing to linguistic distancing across age groups \cite{Nook2017}.
Therefore, we provide new findings that lay the foundation for future exploration into what drives changes in linguistic distancing with age.

\section{Conclusion}

We study how linguistic distancing changes across adulthood to better understand emotion regulation with age. We use large social media datasets to systematically analyze how does linguistic distancing vary across teens to 70's on X and Reddit. We construct an interpretable measure of linguistic distancing based on four dimensions: \textit{temporal} distancing, \textit{social} distancing, \textit{passive} voice and \textit{abstractness}. We found that linguistic distancing occurs in proportionally more posts with age.
% , occurring more frequently in posts. 
When exploring how each of the dimensions changed across age groups, we found that \textit{social} distancing has the highest rate of increase. We also saw that \textit{temporal} distancing plateaus and decreases in later ages. Our work provides an initial foundation for understanding how linguistic distancing varies in text across age groups and supports further exploration into measuring linguistic distancing on social media. Future work could examine how such measures change across platforms discussing diverse topics (e.g., subreddits) and contrast results with LLM measures of linguistic distancing. Future work could also explore linguistic distancing in different languages and cultures, as psychological distancing and distancing in language may differ based on culture or contexts.

% \begin{table}[!ht]
% \begin{center}
% \begin{tabularx}{\columnwidth}{|l|X|}

%       \hline
%       Level&Tools\\
%       \hline
%       Morphology & Pitrat Analyser\\
%       \hline
%       Syntax & LFG Analyser (C-Structure)\\
%       \hline
%      Semantics & LFG F-Structures + Sowa's\\
%      & Conceptual Graphs\\
%       \hline

% \end{tabularx}
% \caption{The caption of the table}
%  \end{center}
% \end{table}

% \section{Acknowledgments}

\section*{Ethical Considerations}
Our research interest is to study how linguistic distancing changes across adulthood at the aggregate/group level. This has applications in emotional development psychology and in public health (e.g., overall well-being and mental health). 
However, personal well-being is complex, private, and central to an individual’s experience. 
Additionally, each individual expresses themselves differently through language, which results in large amounts of variation.
% As any work focusing on well-being and mental health, great care is taken to 

Our work on studying linguistic distancing
should not be construed as detecting how people feel; rather, we draw inferences on the language used.

The inferences we draw in this paper are based
on aggregate trends across large populations.
We do not draw conclusions about specific
individuals or momentary states of well-being.

\section*{Limitations}

The dataset used in this study relies on self-disclosure of age on social media. As with all scenarios involving self-disclosure, individuals may falsely report their age due to social pressures, or in order to relate to and engage with certain communities. Additionally, the templates used for pattern matching with age may not fully capture all possible ways to express age, so there may be age declarations not included in the dataset.

\section*{Acknowledgments}
% This research was supported by the Alberta Machine Intelligence Institute (Amii), the Canadian Institute for Advanced Research (CIFAR), and the Natural Sciences and Engineering Research Council of Canada (NSERC).
This research was supported by the Natural Sciences and Engineering Research Council of Canada (NSERC), the Social Sciences and Humanities Research Council (SSHRC), the Digital Research Alliance of Canada (alliancecan.ca), and the Canadian Institute for Advanced Research (CIFAR). Compute resources were graciously supplied by Prairies DRI and the Digital Research Alliance of Canada. Alona Fyshe holds a Canada CIFAR AI Chair.

\section{Bibliographical References}
\bibliographystyle{lrec2026-natbib}
\bibliography{lrec2026-example}

@article{Nook2017,
	Title = {A linguistic signature of psychological distancing in emotion regulation},
	Author = {Nook, Erik C and Schleider, Jessica L and Somerville, Leah H},
	DOI = {10.1037/xge0000263},
	Number = {3},
	Volume = {146},
	Month = {March},
	Year = {2017},
	Journal = {Journal of experimental psychology. General},
	ISSN = {0096-3445},
	Pages = {337—346},
	URL = {https://doi.org/10.1037/xge0000263},
}

@article{Nook2022,
author = {Erik C. Nook  and Thomas D. Hull  and Matthew K. Nock  and Leah H. Somerville },
title = {Linguistic measures of psychological distance track symptom levels and treatment outcomes in a large set of psychotherapy transcripts},
journal = {Proceedings of the National Academy of Sciences},
volume = {119},
number = {13},
pages = {e2114737119},
year = {2022},
doi = {10.1073/pnas.2114737119},
URL = {https://www.pnas.org/doi/abs/10.1073/pnas.2114737119},
eprint = {https://www.pnas.org/doi/pdf/10.1073/pnas.2114737119}}

@article{Nook2025,
title = "Emotion Regulation is Associated with Increases in Linguistic Measures of Both Psychological Distancing and Abstractness",
author = "Nook, {Erik C.} and Ahn, {Hayoung E.} and Schleider, {Jessica L.} and Somerville, {Leah H.}",
note = "Publisher Copyright: {\textcopyright} The Author(s) 2024.",
year = "2025",
month = mar,
doi = "10.1007/s42761-024-00269-7",
language = "English (US)",
volume = "6",
pages = "63--76",
journal = "Affective Science",
issn = "2662-2041",
publisher = "Springer Nature",
number = "1",
}

@article{passive_voice_distancing,
author = {Eugene Y. Chan and Sam J. Maglio},
title ={The Voice of Cognition: Active and Passive Voice Influence Distance and Construal},
journal = {Personality and Social Psychology Bulletin},
volume = {46},
number = {4},
pages = {547-558},
year = {2020},
doi = {10.1177/0146167219867784},
    note ={PMID: 31390936},
URL = { 
        https://doi.org/10.1177/0146167219867784
},
eprint = { 
        https://doi.org/10.1177/0146167219867784
}
}

@article{aldao2010,
title = {Emotion-regulation strategies across psychopathology: A meta-analytic review},
journal = {Clinical Psychology Review},
volume = {30},
number = {2},
pages = {217-237},
year = {2010},
issn = {0272-7358},
doi = {https://doi.org/10.1016/j.cpr.2009.11.004},
url = {https://www.sciencedirect.com/science/article/pii/S0272735809001597},
author = {Amelia Aldao and Susan Nolen-Hoeksema and Susanne Schweizer}
}

@article{gross1998,
author = {James J. Gross},
title ={The Emerging Field of Emotion Regulation: An Integrative Review},
journal = {Review of General Psychology},
volume = {2},
number = {3},
pages = {271-299},
year = {1998},
doi = {10.1037/1089-2680.2.3.271},
URL = { 
        https://doi.org/10.1037/1089-2680.2.3.271
},
eprint = { 
        https://doi.org/10.1037/1089-2680.2.3.271
}
}

@article{gross2010,
author = {Heather L. Urry and James J. Gross},
title ={Emotion Regulation in Older Age},
journal = {Current Directions in Psychological Science},
volume = {19},
number = {6},
pages = {352-357},
year = {2010},
doi = {10.1177/0963721410388395},
URL = { https://doi.org/10.1177/0963721410388395
},
eprint = { https://doi.org/10.1177/0963721410388395
}
}

@article {Gross1997,
	Title = {Emotion and aging: experience, expression, and control},
	Author = {Gross, JJ and Carstensen, LL and Pasupathi, M and Tsai, J and Skorpen, CG and Hsu, AY},
	DOI = {10.1037//0882-7974.12.4.590},
	Number = {4},
	Volume = {12},
	Month = {December},
	Year = {1997},
	Journal = {Psychology and aging},
	ISSN = {0882-7974},
	Pages = {590—599},
	URL = {https://doi.org/10.1037//0882-7974.12.4.590},
}

@article {Mehlsen2024,
	Title = {The effect of age on emotion regulation patterns in daily life: Findings from an experience sampling study},
	Author = {Mikkelsen, Mai Bjørnskov and O'Toole, Mia Skytte and Elkjaer, Emma and Mehlsen, Mimi},
	DOI = {10.1111/sjop.12970},
	Number = {2},
	Volume = {65},
	Month = {April},
	Year = {2024},
	Journal = {Scandinavian journal of psychology},
	ISSN = {0036-5564},
	Pages = {231—239},
	URL = {https://doi.org/10.1111/sjop.12970},
}

@incollection{Kross2017,
title = {Chapter Two - Self-Distancing: Theory, Research, and Current Directions},
editor = {James M. Olson},
series = {Advances in Experimental Social Psychology},
publisher = {Academic Press},
volume = {55},
pages = {81-136},
year = {2017},
issn = {0065-2601},
doi = {https://doi.org/10.1016/bs.aesp.2016.10.002},
url = {https://www.sciencedirect.com/science/article/pii/S0065260116300338},
author = {E. Kross and O. Ayduk}
}

@book{beck2020cognitive,
  title={Cognitive Behavior Therapy, Third Edition: Basics and Beyond},
  author={Beck, J.S. and Beck, A.T.},
  isbn={9781462544196},
  lccn={2020006172},
  url={https://books.google.ca/books?d=yb_nDwAAQBAJ},
  year={2020},
  publisher={Guilford Publications}
}

@book{linehan1993skills,
  title={Skills training manual for treating borderline personality disorder.},
  author={Linehan, Marsha M},
  year={1993},
  publisher={Guilford press}
}

@article{abstractness2014,
	Title = {Concreteness ratings for 40 thousand generally known English word lemmas},
	Author = {Brysbaert, Marc and Warriner, Amy Beth and Kuperman, Victor},
	DOI = {10.3758/s13428-013-0403-5},
	Number = {3},
	Volume = {46},
	Month = {September},
	Year = {2014},
	Journal = {Behavior research methods},
	ISSN = {1554-351X},
	Pages = {904—911},
	URL = {https://biblio.ugent.be/publication/5774089/file/5774125.pdf},
}

@article{PassivePy,
author = {Sepehri, Amir and Mirshafiee, Mitra Sadat and Markowitz, David M.},
title = {PassivePy: A tool to automatically identify passive voice in big text data},
journal = {Journal of Consumer Psychology},
volume = {33},
number = {4},
pages = {714-727},
doi = {https://doi.org/10.1002/jcpy.1377},
url = {https://myscp.onlinelibrary.wiley.com/doi/abs/10.1002/jcpy.1377},
eprint = {https://myscp.onlinelibrary.wiley.com/doi/pdf/10.1002/jcpy.1377},
year = {2023}
}

@inproceedings{mohammad-2011-upon,
    title = "From Once Upon a Time to Happily Ever After: Tracking Emotions in Novels and Fairy Tales",
    author = "Mohammad, Saif",
    editor = "Zervanou, Kalliopi  and
      Lendvai, Piroska",
    booktitle = "Proceedings of the 5th {ACL}-{HLT} Workshop on Language Technology for Cultural Heritage, Social Sciences, and Humanities",
    month = jun,
    year = "2011",
    address = "Portland, OR, USA",
    publisher = "Association for Computational Linguistics",
    url = "https://aclanthology.org/W11-1514/",
    pages = "105--114"
}

@article{reagan,
author={Reagan,Andrew J. and Mitchell,Lewis and Kiley,Dilan and Danforth,Christopher M. and Dodds,Peter S.},
year={2016},
title={The emotional arcs of stories are dominated by six basic shapes},
journal={EPJ Data Science},
volume={5},
number={1},
pages={31},
language={English},
url={https://login.ezproxy.library.ualberta.ca/login?url=https://www.proquest.com/scholarly-journals/emotional-arcs-stories-are-dominated-six-basic/docview/1865288690/se-2},
}

@article {erq,
	Title = {Individual differences in two emotion regulation processes: implications for affect, relationships, and well-being},
	Author = {Gross, James J and John, Oliver P},
	DOI = {10.1037/0022-3514.85.2.348},
	Number = {2},
	Volume = {85},
	Month = {August},
	Year = {2003},
	Journal = {Journal of personality and social psychology},
	ISSN = {0022-3514},
	Pages = {348—362},
	URL = {https://doi.org/10.1037/0022-3514.85.2.348},
}

@article{Nook2022_adolescent,
title = {Linguistic distancing predicts response to a digital single-session intervention for adolescent depression},
journal = {Behaviour Research and Therapy},
volume = {159},
pages = {104220},
year = {2022},
issn = {0005-7967},
doi = {https://doi.org/10.1016/j.brat.2022.104220},
url = {https://www.sciencedirect.com/science/article/pii/S0005796722001917},
author = {Katherine A. Cohen and Akash Shroff and Erik C. Nook and Jessica L. Schleider}
}

@article {Hu2014,
	Title = {Relation between emotion regulation and mental health: a meta-analysis review},
	Author = {Hu, Tianqiang and Zhang, Dajun and Wang, Jinliang and Mistry, Ritesh and Ran, Guangming and Wang, Xinqiang},
	DOI = {10.2466/03.20.pr0.114k22w4},
	Number = {2},
	Volume = {114},
	Month = {April},
	Year = {2014},
	Journal = {Psychological reports},
	ISSN = {0033-2941},
	Pages = {341—362},
	URL = {https://doi.org/10.2466/03.20.PR0.114k22w4},
}

@article{gross1995,
author = {Gross, James J. and Muñoz, Ricardo F.},
title = {Emotion Regulation and Mental Health},
journal = {Clinical Psychology: Science and Practice},
volume = {2},
number = {2},
pages = {151-164},
doi = {https://doi.org/10.1111/j.1468-2850.1995.tb00036.x},
url = {https://onlinelibrary.wiley.com/doi/abs/10.1111/j.1468-2850.1995.tb00036.x},
eprint = {https://onlinelibrary.wiley.com/doi/pdf/10.1111/j.1468-2850.1995.tb00036.x},
year = {1995}
}

@book{bowlby1969attachment,
  title={Attachment and loss},
  author={Bowlby, John},
  number={79},
  year={1969},
  publisher={New York: Basic Books}
}

@article{Derryberry1988,
	Title = {Arousal, affect, and attention as components of temperament},
	Author = {Derryberry, D and Rothbart, MK},
	DOI = {10.1037//0022-3514.55.6.958},
	Number = {6},
	Volume = {55},
	Month = {December},
	Year = {1988},
	Journal = {Journal of personality and social psychology},
	ISSN = {0022-3514},
	Pages = {958—966},
	URL = {https://doi.org/10.1037//0022-3514.55.6.958},
}

@article{John2004,
author = {John, Oliver P. and Gross, James J.},
title = {Healthy and Unhealthy Emotion Regulation: Personality Processes, Individual Differences, and Life Span Development},
journal = {Journal of Personality},
volume = {72},
number = {6},
pages = {1301-1334},
doi = {https://doi.org/10.1111/j.1467-6494.2004.00298.x},
url = {https://onlinelibrary.wiley.com/doi/abs/10.1111/j.1467-6494.2004.00298.x},
eprint = {https://onlinelibrary.wiley.com/doi/pdf/10.1111/j.1467-6494.2004.00298.x},
year = {2004}
}

@article {DeFrance2019,
	Title = {Emotion regulation and relations to well-being across the lifespan},
	Author = {De France, Kalee and Hollenstein, Tom},
	DOI = {10.1037/dev0000744},
	Number = {8},
	Volume = {55},
	Month = {August},
	Year = {2019},
	Journal = {Developmental psychology},
	ISSN = {0012-1649},
	Pages = {1768—1774},
	URL = {https://doi.org/10.1037/dev0000744},
}

@article{Sutterlin2012,
author = {Sütterlin, Stefan and Paap, Muirne C. S. and Babic, Stana and Kübler, Andrea and Vögele, Claus},
title = {Rumination and Age: Some Things Get Better},
journal = {Journal of Aging Research},
volume = {2012},
number = {1},
pages = {267327},
doi = {https://doi.org/10.1155/2012/267327},
url = {https://onlinelibrary.wiley.com/doi/abs/10.1155/2012/267327},
eprint = {https://onlinelibrary.wiley.com/doi/pdf/10.1155/2012/267327},
year = {2012}
}

@article{Brummer_2014,
title={The Influence of Age on Emotion Regulation Strategies and Psychological Distress},
volume={42}, 
DOI={10.1017/S1352465813000453},
number={6},
journal={Behavioural and Cognitive Psychotherapy},
author={Brummer, Laura and Stopa, Lusia and Bucks, Romola},
year={2014}, 
pages={668–681}
}

@article{Nook2020Age,
title = "Use of linguistic distancing and cognitive reappraisal strategies during emotion regulation in children, adolescents, and young adults",
author = "Nook, {Erik C.} and {Vidal Bustamante}, {Constanza M.} and Cho, {Hyun Young} and Somerville, {Leah H.}",
note = "Publisher Copyright: {\textcopyright} 2020 American Psychological Association.",
year = "2020",
month = jun,
doi = "10.1037/emo0000570",
language = "English (US)",
volume = "20",
pages = "525--540",
journal = "Emotion",
issn = "1528-3542",
publisher = "American Psychological Association",
number = "4",
}

@article {Trope2010,
	Title = {Construal-level theory of psychological distance},
	Author = {Trope, Yaacov and Liberman, Nira},
	DOI = {10.1037/a0018963},
	Number = {2},
	Volume = {117},
	Month = {April},
	Year = {2010},
	Journal = {Psychological review},
	ISSN = {0033-295X},
	Pages = {440—463},
	URL = {https://europepmc.org/articles/PMC3152826},
}

@article {Liberman2008,
	Title = {The psychology of transcending the here and now},
	Author = {Liberman, Nira and Trope, Yaacov},
	DOI = {10.1126/science.1161958},
	Number = {5905},
	Volume = {322},
	Month = {November},
	Year = {2008},
	Journal = {Science (New York, N.Y.)},
	ISSN = {0036-8075},
	Pages = {1201—1205},
	URL = {https://europepmc.org/articles/PMC2643344},
}

@article{Altshuler1995,
title = {Children's knowledge and use of coping strategies during hospitalization for elective surgery},
journal = {Journal of Applied Developmental Psychology},
volume = {16},
number = {1},
pages = {53-76},
year = {1995},
issn = {0193-3973},
doi = {https://doi.org/10.1016/0193-3973(95)90016-0},
url = {https://www.sciencedirect.com/science/article/pii/0193397395900160},
author = {Jennifer L. Altshuler and Janice L. Genevaro and Diane N. Ruble and Marc H. Bonstein}
}

@article{Brown1991,
 ISSN = {0272930X, 15350266},
 URL = {http://www.jstor.org/stable/23087366},
 author = {Kirk Brown and Katherine Covell and Rona Abramovitch},
 journal = {Merrill-Palmer Quarterly},
 number = {2},
 pages = {273--287},
 publisher = {Wayne State University Press},
 title = {Time Course and Control of Emotion: Age Differences in Understanding and Recognition},
 urldate = {2026-02-20},
 volume = {37},
 year = {1991}
}

@article {Stawski2008,
	Title = {Reported exposure and emotional reactivity to daily stressors: the roles of adult age and global perceived stress},
	Author = {Stawski, Robert S and Sliwinski, Martin J and Almeida, David M and Smyth, Joshua M},
	DOI = {10.1037/0882-7974.23.1.52},
	Number = {1},
	Volume = {23},
	Month = {March},
	Year = {2008},
	Journal = {Psychology and aging},
	ISSN = {0882-7974},
	Pages = {52—61},
	URL = {https://europepmc.org/articles/PMC3485068},
}

@article{CERQ,
author = {Garnefski, Nadia and Kraaij, Vivian},
year = {2007},
month = {01},
pages = {141-149},
title = {The Cognitive Emotion Regulation Questionnaire},
volume = {23},
journal = {European Journal of Psychological Assessment},
doi = {10.1027/1015-5759.23.3.141}
}

@article {Sheppes2015,
	Title = {Emotion regulation and psychopathology},
	Author = {Sheppes, Gal and Suri, Gaurav and Gross, James J},
	DOI = {10.1146/annurev-clinpsy-032814-112739},
	Volume = {11},
	Year = {2015},
	Journal = {Annual review of clinical psychology},
	ISSN = {1548-5943},
	Pages = {379—405},
	URL = {https://doi.org/10.1146/annurev-clinpsy-032814-112739},
}

@article{Thompson1991,
 ISSN = {1040726X, 1573336X},
 URL = {http://www.jstor.org/stable/23359228},
 author = {Ross A. Thompson},
 journal = {Educational Psychology Review},
 number = {4},
 pages = {269--307},
 publisher = {Springer},
 title = {Emotional Regulation and Emotional Development},
 urldate = {2026-02-20},
 volume = {3},
 year = {1991}
}

@article{hall1993nicotine,
  title={Nicotine, negative affect, and depression.},
  author={Hall, Sharon M and Mu{\~n}oz, Ricardo F and Reus, Victor I and Sees, Karen L},
  journal={Journal of consulting and clinical psychology},
  volume={61},
  number={5},
  pages={761},
  year={1993},
  publisher={American Psychological Association}
}

@article{Felix1995,
author = {María Félix-Ortiz and Ricardo Muñoz and Michael D. Newcomb},
title = {The Role of Emotional Distress in Drug Use Among Latino Adolescents},
journal = {Journal of Child \& Adolescent Substance Abuse},
volume = {3},
number = {4},
pages = {1--22},
year = {1995},
publisher = {Taylor \& Francis},
doi = {10.1300/J029v03n04\_01},
URL = {    https://doi.org/10.1300/J029v03n04_01
},
eprint = {  
        https://doi.org/10.1300/J029v03n04_01
}
}

@article{eugene2020,
author = {Eugene Y. Chan and Sam J. Maglio},
title ={The Voice of Cognition: Active and Passive Voice Influence Distance and Construal},
journal = {Personality and Social Psychology Bulletin},
volume = {46},
number = {4},
pages = {547-558},
year = {2020},
doi = {10.1177/0146167219867784},
    note ={PMID: 31390936},
URL = { 
        https://doi.org/10.1177/0146167219867784
},
eprint = {     
        https://doi.org/10.1177/0146167219867784
}
}

@article {Trope2003,
	Title = {Temporal construal},
	Author = {Trope, Yaacov and Liberman, Nira},
	DOI = {10.1037/0033-295x.110.3.403},
	Number = {3},
	Volume = {110},
	Month = {July},
	Year = {2003},
	Journal = {Psychological review},
	ISSN = {0033-295X},
	Pages = {403—421},
	URL = {https://doi.org/10.1037/0033-295x.110.3.403},
}

@misc{teodorescu2026_age,
      title={Age and Affect in Language: How Emotion Expression on Social Media Varies Across Adulthood}, 
      author={Daniela Teodorescu and Jan Wahle and Saif M. Mohammad},
      year={2026},
    booktitle = "Proceedings of the 1st Workshop on Computational Affective Science (CAS 2026)",
    month = may,
    year = "2026",
    address = "Palma de Mallorca, Spain",
    publisher = "European Language Resources Association (ELRA)",
}

@article{Bohner2001,
author = {Bohner, Gerd},
title = {Writing about rape: Use of the passive voice and other distancing text features as an expression of perceived responsibility of the victim},
journal = {British Journal of Social Psychology},
volume = {40},
number = {4},
pages = {515-529},
doi = {https://doi.org/10.1348/014466601164957},
url = {https://bpspsychub.onlinelibrary.wiley.com/doi/abs/10.1348/014466601164957},
eprint = {https://bpspsychub.onlinelibrary.wiley.com/doi/pdf/10.1348/014466601164957},
year = {2001}
}

@article {Charles2023,
	Title = {Growing old and being old: Emotional well-being across adulthood},
	Author = {Charles, Susan T and Rush, Jonathan and Piazza, Jennifer R and Cerino, Eric S and Mogle, Jaqueline and Almeida, David M},
	DOI = {10.1037/pspp0000453},
	Number = {2},
	Volume = {125},
	Month = {August},
	Year = {2023},
	Journal = {Journal of personality and social psychology},
	ISSN = {0022-3514},
	Pages = {455—469},
	URL = {https://europepmc.org/articles/PMC10330366},
}

@article{Stone2010,
author = {Arthur A. Stone  and Joseph E. Schwartz  and Joan E. Broderick  and Angus Deaton },
title = {A snapshot of the age distribution of psychological well-being in the United States},
journal = {Proceedings of the National Academy of Sciences},
volume = {107},
number = {22},
pages = {9985-9990},
year = {2010},
doi = {10.1073/pnas.1003744107},
URL = {https://www.pnas.org/doi/abs/10.1073/pnas.1003744107},
eprint = {https://www.pnas.org/doi/pdf/10.1073/pnas.1003744107}}

@article {Isaacowitz,
	Title = {What Do We Know About Aging and Emotion Regulation?},
	Author = {Isaacowitz, Derek M},
	DOI = {10.1177/17456916211059819},
	Number = {6},
	Volume = {17},
	Month = {November},
	Year = {2022},
	Journal = {Perspectives on psychological science : a journal of the Association for Psychological Science},
	ISSN = {1745-6916},
	Pages = {1541—1555},
	URL = {https://europepmc.org/articles/PMC9633333},
}

@article {Blanchflower2021,
	Title = {The U Shape of Happiness: A Response},
	Author = {Blanchflower, David G and Graham, Carol L},
	DOI = {10.1177/1745691620984393},
	Number = {6},
	Volume = {16},
	Month = {November},
	Year = {2021},
	Journal = {Perspectives on psychological science : a journal of the Association for Psychological Science},
	ISSN = {1745-6916},
	Pages = {1435—1446},
	URL = {https://doi.org/10.1177/1745691620984393},
}

@article{Blanchflower2008,
title = {Is well-being U-shaped over the life cycle?},
journal = {Social Science \& Medicine},
volume = {66},
number = {8},
pages = {1733-1749},
year = {2008},
issn = {0277-9536},
doi = {https://doi.org/10.1016/j.socscimed.2008.01.030},
url = {https://www.sciencedirect.com/science/article/pii/S0277953608000245},
author = {David G. Blanchflower and Andrew J. Oswald}
}

@article {Galambos,
	Title = {The U Shape of Happiness Across the Life Course: Expanding the Discussion},
	Author = {Galambos, Nancy L and Krahn, Harvey J and Johnson, Matthew D and Lachman, Margie E},
	DOI = {10.1177/1745691620902428},
	Number = {4},
	Volume = {15},
	Month = {July},
	Year = {2020},
	Journal = {Perspectives on psychological science : a journal of the Association for Psychological Science},
	ISSN = {1745-6916},
	Pages = {898—912},
	URL = {https://europepmc.org/articles/PMC7529452},
}

@article{Baldie1976, 
title={The acquisition of the passive voice}, 
volume={3}, 
DOI={10.1017/S0305000900007224}, number={3},
journal={Journal of Child Language},
author={Baldie, Brian J.},
year={1976},
pages={331–348}}

@article{Horgan_1978,
title={The development of the full passive}, 
volume={5},
DOI={10.1017/S030500090000194X},
number={1},
journal={Journal of Child Language},
author={Horgan, Dianne}, 
year={1978}, 
pages={65–80}}

@article{Schwartz2013,
    doi = {10.1371/journal.pone.0073791},
    author = {Schwartz, H. Andrew AND Eichstaedt, Johannes C. AND Kern, Margaret L. AND Dziurzynski, Lukasz AND Ramones, Stephanie M. AND Agrawal, Megha AND Shah, Achal AND Kosinski, Michal AND Stillwell, David AND Seligman, Martin E. P. AND Ungar, Lyle H.},
    journal = {PLOS ONE},
    publisher = {Public Library of Science},
    title = {Personality, Gender, and Age in the Language of Social Media: The Open-Vocabulary Approach},
    year = {2013},
    month = {09},
    volume = {8},
    url = {https://doi.org/10.1371/journal.pone.0073791},
    pages = {1-16},
    number = {9},

}

@article{Pang2025,
author={Pang,Dandan and Guntuku,Sharath C. and Sherman,Garrick and Liu,Tingting and Rai,Sunny and Cho,Young-Min and You,Rian and Bakker,Arnold B. and Ungar,Lyle H.},
year={2025},
month={12},
title={Understanding gender and age differences in language use: cross-cultural insights from Weibo and Facebook},
journal={Humanities \& Social Sciences Communications},
volume={12},
number={1},
pages={1667},
note={Copyright - © The Author(s) 2025. This work is published under http://creativecommons.org/licenses/by-nc-nd/4.0/ (the “License”). Notwithstanding the ProQuest Terms and Conditions, you may use this content in accordance with the terms of the License; Last updated - 2025-11-07; SubjectsTermNotLitGenreText - United States--US; China},
language={English},
url={https://login.ezproxy.library.ualberta.ca/login?url=https://www.proquest.com/scholarly-journals/understanding-gender-age-differences-language-use/docview/3267562150/se-2},
}

@article {Pennebaker2003,
	Title = {Words of wisdom: language use over the life span},
	Author = {Pennebaker, James W and Stone, Lori D},
	DOI = {10.1037/0022-3514.85.2.291},
	Number = {2},
	Volume = {85},
	Month = {August},
	Year = {2003},
	Journal = {Journal of personality and social psychology},
	ISSN = {0022-3514},
	Pages = {291—301},
	URL = {https://doi.org/10.1037/0022-3514.85.2.291},
}

@article{Jee2024,
author = {Jee Eun Sung  and Eunha Jo  and Sujin Choi  and Jiyeon Lee },
title = {Coordinating Words and Sentences: Detecting Age-Related Changes in Language Production},
journal = {Journal of Speech, Language, and Hearing Research},
volume = {67},
number = {1},
pages  = {211-220},
year = {2024},
doi = {10.1044/2023\_JSLHR-23-00222},
URL = {https://pubs.asha.org/doi/abs/10.1044/2023_JSLHR-23-00222},
eprint = {https://pubs.asha.org/doi/pdf/10.1044/2023_JSLHR-23-00222}
}

@article {Albert1990,
	Title = {Differences in abstraction ability with age},
	Author = {Albert, MS and Wolfe, J and Lafleche, G},
	DOI = {10.1037//0882-7974.5.1.94},
	Number = {1},
	Volume = {5},
	Month = {March},
	Year = {1990},
	Journal = {Psychology and aging},
	ISSN = {0882-7974},
	Pages = {94—100},
	URL = {https://doi.org/10.1037//0882-7974.5.1.94},
}

@article{pennebaker2001linguistic,
  title={Linguistic inquiry and word count: LIWC 2001},
  author={Pennebaker, James W and others}
}

@inproceedings{ireland-etal-2022-tracking,
    title = "Tracking Mental Health Risks and Coping Strategies in Healthcare Workers' Online Conversations Across the {COVID}-19 Pandemic",
    author = "Ireland, Molly  and
      Adams, Kaitlin  and
      Farrell, Sean",
    editor = "Zirikly, Ayah  and
      Atzil-Slonim, Dana  and
      Liakata, Maria  and
      Bedrick, Steven  and
      Desmet, Bart  and
      Ireland, Molly  and
      Lee, Andrew  and
      MacAvaney, Sean  and
      Purver, Matthew  and
      Resnik, Rebecca  and
      Yates, Andrew",
    booktitle = "Proceedings of the Eighth Workshop on Computational Linguistics and Clinical Psychology",
    month = jul,
    year = "2022",
    address = "Seattle, USA",
    publisher = "Association for Computational Linguistics",
    url = "https://aclanthology.org/2022.clpsych-1.7/",
    doi = "10.18653/v1/2022.clpsych-1.7",
    pages = "76--88"
}

@inproceedings{troiano-etal-2024-dealing,
    title = "Dealing with Controversy: An Emotion and Coping Strategy Corpus Based on Role Playing",
    author = "Troiano, Enrica  and
      Labat, Sofie  and
      Stranisci, Marco Antonio  and
      Damiano, Rossana  and
      Patti, Viviana  and
      Klinger, Roman",
    editor = "Al-Onaizan, Yaser  and
      Bansal, Mohit  and
      Chen, Yun-Nung",
    booktitle = "Findings of the Association for Computational Linguistics: EMNLP 2024",
    month = nov,
    year = "2024",
    address = "Miami, Florida, USA",
    publisher = "Association for Computational Linguistics",
    url = "https://aclanthology.org/2024.findings-emnlp.89/",
    doi = "10.18653/v1/2024.findings-emnlp.89",
    pages = "1634--1658"
}

@article{Pezzuti,
author = {Lina Pezzuti and James Dawe and Anna Maria Borghi},
title = {Does mastering of abstract words decline with age?},
journal = {Educational Gerontology},
volume = {47},
number = {12},
pages = {527--542},
year = {2021},
publisher = {Routledge},
doi = {10.1080/03601277.2021.2008709},
URL = { 
https://doi.org/10.1080/03601277.2021.2008709
},
eprint = { https://doi.org/10.1080/03601277.2021.2008709
}
}

\bibliographystylelanguageresource{lrec2026-natbib}
\bibliographylanguageresource{languageresource}

\appendix

\section{Patterns to Match Age Declarations}
\label{appendix:regexes}

In Table \ref{tab:regexes}, we show the pattern templates used to match age declarations in seed posts in the \textit{AgeCorpus}.

\begin{table*}
  \centering
  \small
  \begin{tabular}{ll}
    \hline
    \textbf{Regex} & \textbf{Example} \\\hline %\textbf{Source} 
% `$\textbackslash$bI(?:$\backslash$s+am|'m)$\textbackslash$s+($\textbackslash$d\{1,2\})$\textbackslash$s+years?$\textbackslash$s+old$\textbackslash$b'
\makecell[l]{
\texttt{\textbackslash bI(?:\textbackslash s+am|'m)\textbackslash s+(\textbackslash d\{1,2\})\textbackslash s+years?\textbackslash s+old\textbackslash b}
} & \makecell[l]{\textit{I am 25 years old} \\ \textit{I'm 30 year old}} \\
    % &  & \\
    \hline
    % & & \\
    
    % \makecell[l]{`$\textbackslash$bI(?:$\textbackslash$s+am|'m)$\textbackslash$s+($\textbackslash$d\{1,2\}) \\ (?=$\textbackslash$s*(?:\$|[,.!?;:$\textbackslash$-]|(?:and|but|so|yet)$\textbackslash$s))'}
    \makecell[l]{\texttt{\textbackslash bI(?:\textbackslash s+am|'m)\textbackslash s+(\textbackslash d\{1,2\})}\\ \texttt{(?=\textbackslash s*(?:\$|[,.!?;:\textbackslash -]|(?:and|but|so|yet)\textbackslash s))}}    &  \makecell[l]{\textit{I am 25.} \\ \textit{I'm 30, and ...}} \\
    \hline 
    % &   &\\ 
    
    % \makecell[l]{`$\textbackslash$bI(?:$\textbackslash$s+was|$\textbackslash$s+am|'m)$\textbackslash$s+born$\textbackslash$s+in$\textbackslash$s+ \\ (19$\textbackslash$d\{2\}|20(?:0$\textbackslash$d|1$\textbackslash$d|2[0-4]))$\textbackslash$b`} 
    \makecell[l]{\texttt{\textbackslash bI(?:\textbackslash s+was|\textbackslash s+am|'m)\textbackslash s+born\textbackslash s+in\textbackslash s+}\\ \texttt{(19\textbackslash d\{2\}|20(?:0\textbackslash d|1\textbackslash d|2[0-4]))\textbackslash b}}    & \makecell[l]{\textit{I was born in 1998} \\ \textit{I am born in 2005}} \\\hline
    % `$\textbackslash$bI(?:$\textbackslash$s+was|$\textbackslash$s+am|'m)$\textbackslash$s+born$\textbackslash$s+in$\textbackslash$s+'($\textbackslash$d\{2\})$\textbackslash$b'
    \makecell[l]{\texttt{\textbackslash bI(?:\textbackslash s+was|\textbackslash s+am|'m)\textbackslash s+born\textbackslash s+in\textbackslash s+'(\textbackslash d\{2\})\textbackslash b}}    & \textit{I was born in '98} \\
    & \textit{I'm born in '05} \\\hline

    \makecell[l]{
\texttt{\textbackslash bI\textbackslash s+was\textbackslash s+born\textbackslash s+on\textbackslash s+}\\
\texttt{(?:\textbackslash d\{1,2\}(?:st|nd|rd|th)?\textbackslash s+)?}\\
\texttt{(?:January|February|March|April|May|June|}\\
\texttt{July|August|September|October|November|December|}\\
\texttt{Jan|Feb|Mar|Apr|May|Jun|Jul|Aug|Sep|Sept|Oct|Nov|Dec)}\\\texttt{\textbackslash s+}
\texttt{(?:\textbackslash d\{1,2\}(?:st|nd|rd|th)?,?\textbackslash s+)?}\\
\texttt{(19\textbackslash d\{2\}|20(?:0\textbackslash d|1\textbackslash d|2[0-4]))\textbackslash b}
} & \makecell[l]{\textit{I was born on 15 March 1998} \\  \textit{I was born on March 15th, 1998}} \\\hline

    % \makecell[l]{`$\textbackslash$bI$\textbackslash$s+was$\textbackslash$s+born$\textbackslash$s+on$\textbackslash$s+$\textbackslash$d\{1,2\}[\slash$\textbackslash$-]$\textbackslash$d\{1,2\} \\ [\slash$\textbackslash$-](19$\textbackslash$d\{2\}|20(?:0$\textbackslash$d|1$\textbackslash$d|2[0-4]))$\textbackslash$b`}
  \makecell[l]{\texttt{\textbackslash bI\textbackslash s+was\textbackslash s+born\textbackslash s+on\textbackslash s+}\textbackslash \texttt{d\{1,2\}} \\ \texttt{[/\textbackslash-]\textbackslash d\{1,2\}[/\textbackslash-](19\textbackslash d\{2\}|20(?:0\textbackslash d|1\textbackslash d|2[0-4]))\textbackslash b}} & \makecell[l]{\textit{I was born on 03\slash15\slash1998} \\ \textit{I was born on 15-03-1998}} \\\hline
    
    \hline
  \end{tabular}
  \caption{Regexes used to identify users from both the Reddit and X datasets.}
  \label{tab:regexes}
\end{table*}

\section{Levene's Test of Homogeneity of Variance}
\label{appendix:levenes}
Levene’s test indicated that the assumption for homogeneity of variance was not violated for the effect of age group on 
linguistic distancing for TUSC-Country, but was violated for Reddit and TUSC-City 
% all metrics (linguistic distance, temporal distance, social distance, passive voice and abstractness) 
subsets of the \textit{AgeCorpus}. We show the results in Table \ref{tab:levens}.

\begin{table*}[!ht]
\centering
{\small
\begin{tabular}
{llrrrr}
\hline
\multicolumn{1}{l}{\textbf{Dataset}} &
\multicolumn{1}{l}{\textbf{Metric}} &
\multicolumn{1}{l}{\textbf{df1}} &
\multicolumn{1}{l}{\textbf{df2}} &
\multicolumn{1}{l}{\textbf{F-statistic}} &
\multicolumn{1}{l}{\textbf{P-value}} \\ \hline

Reddit & Linguistic & 6 & 32964822 & 10269.65 & \textit{p}$<$.001 \\
% & Temporal & 6 &  & 32941.477 & \textit{p}$<$.001\\
% & Social & 6 &  &  & \textit{p}$<$.001\\
% & Passive & 6 &  & & \textit{p}$<$.001 \\ 
% & Abstract & 6 &  &  & \textit{p}$<$.001 \\ 

TUSC-City & Linguistic & 6 & 1918310 & 86.13 & \textit{p}$<$.001 \\
% & Temporal & 6 &  &  & \textit{p}$<$.001\\
% & Social & 6 &  &  & \textit{p}$<$.001\\
% & Passive & 6 &  & & \textit{p}$<$.001 \\ 
% & Abstract & 6 &  &  & \textit{p}$<$.001 \\ 
 
TUSC-Country & Linguistic & 6 & 11,940 & 1.12 & \textit{p}$=$.348 \\
% & Temporal & 6 & 11,940 & 4.00 & \textit{p}$<$.001\\
% & Social & 6 & 11,940 & 4.80  & \textit{p}$<$.001\\
% & Passive & 6 & 11,940 & 7.71 & \textit{p}$<$.001 \\ 
% & Abstract & 6 & 11,940 & 3.20 & \textit{p}$=$.004 \\ 

\hline
\end{tabular}
}
% \vspace*{-1mm}
\caption{The degrees of freedom, F-statistic, and p-value in Levene's test of Homogeneity of Variances for linguistic distancing and each dimension of linguistic distancing across the subsets of the \textit{AgeCorpus}.}
\vspace*{3mm}
\label{tab:levens}
\end{table*}

\begin{table}[t]
\centering
\begin{tabular}{lrrr}
\hline
\textbf{Age Group} & \multicolumn{3}{c}{\textbf{\#Posts}}\\
\cline{2-4}
 & Country  & City & Reddit\\ 
\hline
13--19 & 551 & 94,857 & 9,281,055\\
20--29 & 3,076 & 406,027 & 15,455,426\\
30--39 & 3,001 & 462,238 & 6,151,757\\
40--49 & 1,808 & 360,946 & 1,220,498\\
50--59 & 1,541 & 278,809 & 449,230\\
60--69 & 1,551 & 218,008 & 246,398\\
70--79 & 419 & 97,432 & 160,465\\
\hline
\end{tabular}
\caption{The number of posts across the age groups in each subset of the \textit{AgeCorpus}. ``Country'' refers to TUSC-Country and ``City'' refers to TUSC-City subsets of the dataset. We use this data in our experiments.
}
\label{tab:AgeCorpus2}
\end{table}
\section{TUSC-Country Results}
\label{appendix:tusc_country}

In the following Sections we show the number of posts in the TUSC-Country version of the dataset and the linguistic distancing results on this subset of the \textit{AgeCorpus}.

\subsection{\textit{AgeCorpus}: TUSC-Country }
In Table \ref{tab:AgeCorpus2}, we show the number of posts in TUSC-Country, TUSC-City and Reddit subsections of the \textit{AgeCorpus} for comparison.

\subsection{Linguistic Distancing Results}

In Table \ref{tab:anova_distancing2}, we show the results from the ANOVA tests for the difference in linguistic distancing across age groups. We include the rows for the Reddit and TUSC-City subsets of the \textit{AgeCorpus} dataset as a comparison.

\begin{figure*}[!ht]
    \centering
    % \begin{subfigure}[b]{0.45\textwidth}
    % \centering
    % % \subcaption{}
    % \includegraphics[width=\textwidth]{images/linguistic_distancing_age_city_lineplot.png}
    % % \caption{}
    % \label{fig:tusc_city_linguistic_dist_age}
    % \end{subfigure}
    % \hfill
    \begin{subfigure}[b]{0.45\textwidth}
    \centering
    % \subcaption{}
    \includegraphics[width=\textwidth]{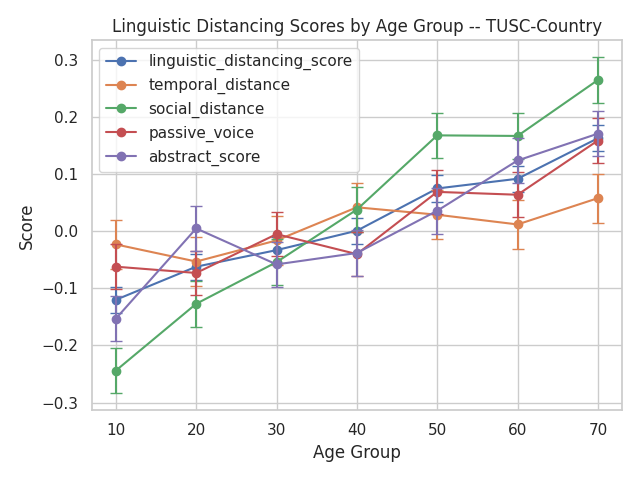}
    % \caption{}
    \label{fig:tusc_country_linguistic_dist_age}
    \end{subfigure}
        \vspace*{-3mm}

    % \begin{subfigure}[b]{0.45\textwidth}
    % \centering
    % % \subcaption{}
    % \includegraphics[width=\textwidth]{images/linguistic_distancing_age_reddit_lineplot.png}
    % % \caption{}
    % \label{fig:tusc_country_linguistic_dist_age}
    % \end{subfigure}
    %     \vspace*{-3mm}        
    \caption{\textbf{Linguistic Distancing} scores for TUSC-Country subset of the \textit{AgeCorpus} (blue line). Individual components of linguistic distancing across age groups are shown: temporal distance (orange line), social distance (green line), passive voice (red line) and abstractness (purple line). Error bars represent the standard error of the mean.}
     
    \label{fig:tusc_linguistic_distancing_age2}
\end{figure*}

In Figure \ref{fig:tusc_linguistic_distancing_age2}, we show the trends of linguistic distancing across age groups (blue line), as well as the trends for \textit{temporal} distancing, \textit{social} distancing, \textit{passive voice} and \textit{abstractness}. Largely the trends for TUSC-Country follow those of TUSC-City, with an increasing proportion of distancing occurring with age. Generally, all dimensions of distancing grow similarly across age groups, with social distancing having the steepest slope.

\begin{table*}[!ht]
\centering
{\small
\begin{tabular}
{llrrrrr}
\hline
\multicolumn{1}{l}{\textbf{Dataset}} &
\multicolumn{1}{l}{\textbf{Metric}} &
\multicolumn{1}{l}{\textbf{df1}} &
\multicolumn{1}{l}{\textbf{df2}} &
\multicolumn{1}{l}{\textbf{F-statistic}} &
\multicolumn{1}{l}{\textbf{P-value}} &
\multicolumn{1}{l}{\textbf{Effect Size (\textit{est $\omega$}\textsuperscript{2})}}\\ \hline
%One-way
% Reddit & Linguistic & 6 & 32964822 & 18001.10 & \textit{p}$<$.001 & 0.003 \\
%Welch's anova
Reddit & Linguistic & 6 & 1203317.10 & 17703.63 & \textit{p}$<$.001 & 0.003 \\
% & Temporal &  & &  & \textit{p}$<$.001 & \\
% & Social &  & &  & \textit{p}$<$.001 &  \\
% & Passive &  &  &  & \textit{p}$<$.001 &  \\
% & Abstract &  &  &  & \textit{p}$<$.001 &  \\

TUSC-Country & Linguistic & 6 & 11940 & 29.66 & \textit{p}$<$.001 & 0.015 \\
% & Temporal & 6 & 11940 & 2.59 & \textit{p}$=$.016 & 0.001\\
% & Social & 6 & 11940 & 35.43 & \textit{p}$<$.001 & 0.017 \\
% & Passive & 6 & 11940 & 7.71 & \textit{p}$<$.001 & 0.004 \\
% & Abstract & 6 & 11940 & 10.68 & \textit{p}$<$.001 & 0.005 \\

% One-way
% TUSC-City & Linguistic & 6 & 1918310 & 2822.47 & \textit{p}$<$.001 & 0.009 \\

% Welch's anova
TUSC-City & Linguistic & 6 & 539267.68 & 2830.26 & \textit{p}$<$.001 & 0.009 \\
% & Temporal & 6 & 1918310 & 218.15 & \textit{p}$<$.001 & 0.001\\
% & Social & 6 & 1918310 & 3636.48 & \textit{p}$<$.001 & 0.011 \\
% & Passive & 6 & 1918310 & 567.58 & \textit{p}$<$.001 & 0.002 \\
% & Abstract & 6 & 1918310 & 853.84 & \textit{p}$<$.001 & 0.003 \\
 
\hline
\end{tabular}
}
% \vspace*{-1mm}
\caption{The degrees of freedom (for the numerator and denominator), F-statistic, p-value, and effect size in the one-way ANOVA test for differences in linguistic distancing between age groups. Welch's ANOVA was performed for Reddit and TUSC-City. % ``emo.'' is an abbreviation for emotional.
}
\vspace*{3mm}
\label{tab:anova_distancing2}
\end{table*}

\end{document}